\documentclass[letterpaper, 10 pt, conference]{ieeeconf}  % Comment this line out if you need a4paper

\IEEEoverridecommandlockouts                              % This command is only needed if
\usepackage{amssymb}  % assumes amsmath package installed
\usepackage[usenames,dvipsnames,table]{xcolor}

\usepackage{booktabs}
\usepackage{colortbl}
\usepackage[iso,german]{isodate}
\usepackage{multirow}
\usepackage{lipsum}
\usepackage{mwe}
\usepackage{pifont}
\usepackage{threeparttable}
\usepackage{tikz}
\newcommand{\fig}[1]{Fig.~\ref{#1}}

\newcommand{\tab}[1]{Table~\ref{#1}}

\newcommand{\sect}[1]{Section~\ref{#1}}

\newcommand{\cmark}{\textcolor{ForestGreen}{\ding{51}}}%
\newcommand{\xmark}{\textcolor{Red}{\ding{55}}}%

\makeatletter
\let\NAT@parse\undefined
\makeatother
\usepackage{hyperref}

\newcommand\copyrighttext{%
    \footnotesize \copyright{ }2024 IEEE. Personal use of this material is permitted. Permission from IEEE must be obtained for all other uses, in any current or future media, including reprinting/republishing this material for advertising or promotional purposes, creating new collective works, for resale or redistribution to servers or lists, or reuse of any copyrighted component of this work in other works.}
\newcommand\copyrightnotice{%
    \AtBeginShipoutNext{\AtBeginShipoutUpperLeftForeground{
        \begin{tikzpicture}[remember picture,overlay]
            \node[anchor=south,yshift=15pt,xshift=0pt] at (current page.south) {\parbox{\dimexpr\textwidth-\fboxsep-\fboxrule\relax}{\copyrighttext}};
        \end{tikzpicture}%
    }}
}

\newcommand\httpsurl[1]{%
  \href{https://#1}{\nolinkurl{#1}}%
}

\title{\LARGE \bf
Enabling the Deployment of Any-Scale Robotic Applications \\in Microservice Architectures through Automated Containerization*
}

\author{Jean-Pierre Busch$^{\dagger}$ and Lennart Reiher$^{\dagger}$, Lutz Eckstein$^{\ddagger}$% <-this % stops a space
  \thanks{*This work is accomplished within the projects 6GEM~(FKZ~16KISK036K) and AUTOtech.\textit{agil}~(FKZ~01IS22088A). We acknowledge the financial support for the projects by the Federal Ministry of Education and Research of Germany~(BMBF).}% <-this % stops a space
  \thanks{$^{\dagger}$The authors contributed equally to this work. They are with the Institute for Automotive Engineering~(ika), RWTH Aachen University, Germany. {\tt\small \{firstname.lastname\}@ika.rwth-aachen.de}}
  \thanks{$^{\ddagger}$Lutz Eckstein is head of the Institute for Automotive Engineering~(ika).}
}

\begin{document}

\bstctlcite{IEEEexample:BSTcontrol}
\maketitle
\thispagestyle{empty}
\pagestyle{empty}

\copyrightnotice

%-------------------------------------------------------------------------------

\begin{abstract}

In an increasingly automated world -- from warehouse robots to self-driving cars -- streamlining the development and deployment process and operations of robotic applications becomes ever more important. Automated DevOps processes and microservice architectures have already proven successful in other domains such as large-scale customer-oriented web services (e.g., Netflix). We recommend to employ similar microservice architectures for the deployment of small- to large-scale robotic applications in order to accelerate development cycles, loosen functional dependence, and improve resiliency and elasticity. In order to facilitate involved DevOps processes, we present and release a tooling suite for automating the development of microservices for robotic applications based on the Robot Operating System~(ROS). Our tooling suite covers the automated minimal containerization of ROS applications, a collection of useful machine learning-enabled base container images, as well as a CLI tool for simplified interaction with container images during the development phase. Within the scope of this paper, we embed our tooling suite into the overall context of streamlined robotics deployment and compare it to alternative solutions. We release our tools as open-source software at \httpsurl{github.com/ika-rwth-aachen/dorotos}.

\end{abstract}

\section{Introduction}
\label{sec:introduction}

Striving for automation of manual labor through robots, be it simple or complex tasks, is as prevalent as ever. Robotic applications span all major domains, from logistics and manufacturing to transportation and healthcare. One of the most booming fields of robotics is the large-scale automation and deployment of connected and intelligent vehicles. Future intelligent transportation systems~(ITS) cover not only automated vehicles, but also include roadside infrastructure and connected cloud- and edge-cloud servers. Such large-scale systems highlight the challenges in deploying modern robotic applications: individual functions (e.g., environment perception) by themselves are already highly complex; applications may underlie real-time constraints; meeting such constraints can be difficult, particularly on resource-constrained hardware; the heterogeneity of involved hardware and software platforms makes it difficult to ensure correct functionality across platforms; (over-the-air)~updates to the system should roll out automatically and independently without downtime; and the distributed nature of running multiple software components across different devices poses many more challenges in deployment and operation.

\begin{figure}[t]
    \centering
    \smallskip
    \smallskip
    \includegraphics[width=\linewidth]{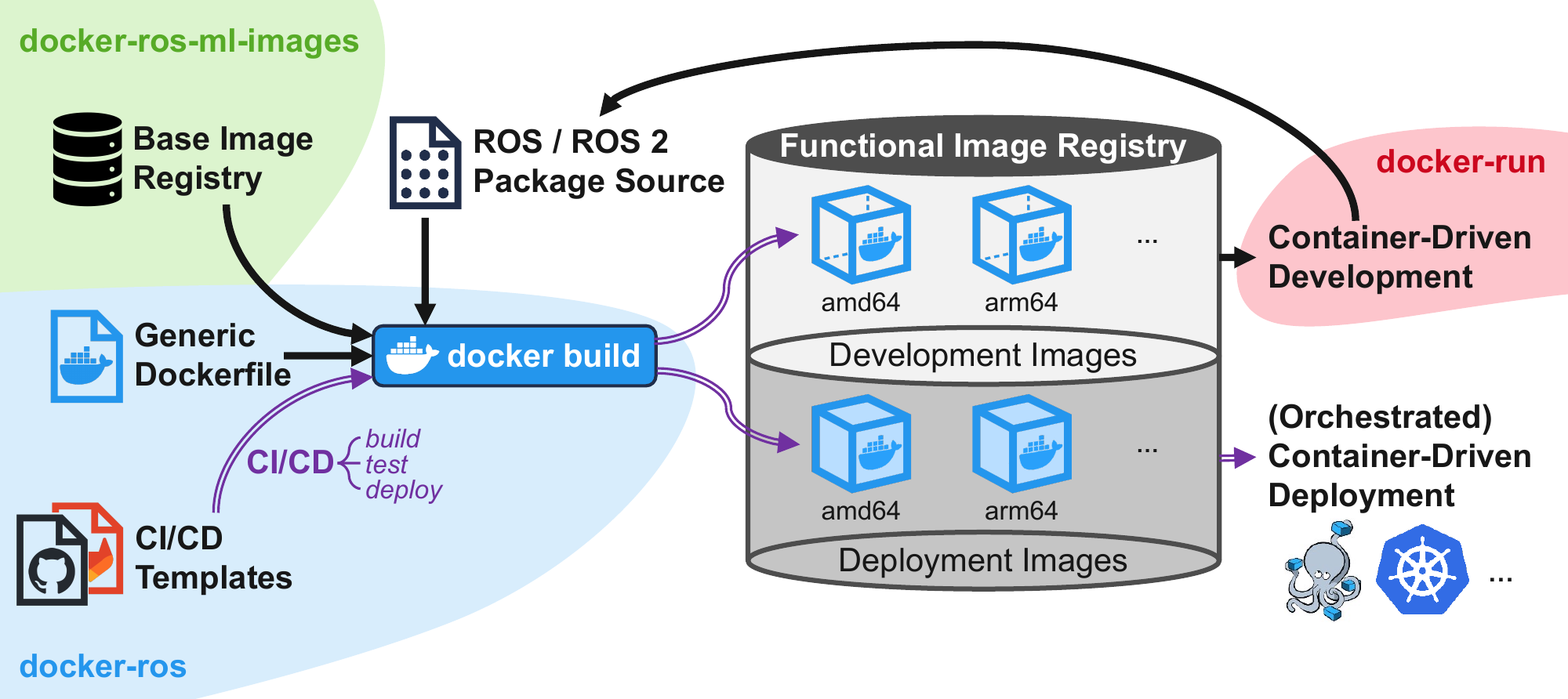}
    \setlength{\abovecaptionskip}{-7pt}
    \caption{Overview of the proposed microservice-based ROS development and deployment workflow using our tooling suite: \textit{docker-ros} provides a generic Dockerfile; integrated via CI/CD templates for popular Git platforms, the container image build is automated for any ROS or ROS~2 package source; two images are built: the development image contains all required dependencies and the source code, the deployment image only contains dependencies and compiled binaries; any base image can be used, but \textit{docker-ros-ml-images} hosts a set of useful machine learning-enabled base images; \textit{docker-run} is a CLI tool to speed up the container-driven development workflow; the final deployment images are designed to be deployed, either standalone or orchestrated by tools like \textit{Kubernetes}.}
    \label{fig:system-architecture}
\end{figure}

Other domains such as large-scale customer-oriented web services (e.g, social media and video streaming) are facing similar challenges in deployment. Not only recently have virtualization/containerization, microservice architectures, and automated DevOps processes proven successful in overcoming these challenges in said domains~\cite{OffermanEtAl_StudyAdoptionEffects_2022}. Containerization, as a form of virtualization, packages only a minimal set of binaries, libraries, and other dependencies into a container image. Lower software layers such as the operating system are shared between isolated containers, thereby reducing overhead. This is beneficial for implementing a microservice architecture, where a monolithic application is broken down into individual, independent, and task-specific software components. The automation of the software development life cycle (from building to testing to release) through DevOps enables quick iterations and scalability and integrates well with containerized microservice architectures. Tools like \textit{Docker}~\cite{Merkel2014DockerLL} and \textit{Kubernetes}~\cite{CloudNativeComputingFoundation_Kubernetes_} have emerged as de facto standards for containerization and the subsequent orchestration of said containers, i.e., automating deployment, scaling, and operation.

% ==============================================================================
\subsection{Deployment of Robotic Applications}
\label{sec:deployment-of-robotic-applications}

In robotics, the \textit{Robot Operating System~(ROS)}~\cite{QuigleyEtAl_ROSOpensourceRobot_2009a, MacenskiEtAl_RobotOperatingSystem_2022} is a widely used open-source middleware suite providing software frameworks, libraries, and tools. ROS follows a modular approach, i.e., it is designed to isolate individual robotic functions into isolated and task-specific so-called ROS nodes. These nodes communicate via clearly defined interfaces and as such can be understood to follow a service-oriented approach. Although ROS loosely integrates DevOps workflow components such as testing or coordinated execution (through \texttt{roslaunch}), ROS on its own is not designed for large-scale interconnected deployments across many robot agents. It lacks required features that higher-level frameworks like Docker and Kubernetes bring, such as isolated environments, proper orchestration, life cycle management, and scaled deployment.

The automated large-scale deployment of robotic applications is tightly bound to the field of cloud robotics. Existing research spans from high-level architecture design to implementing concrete cloud robotics applications. An overview of related works can be found in several survey papers~\cite{HuEtAl_CloudRoboticsArchitecture_2012, KehoeEtAl_SurveyResearchCloud_2015, WanEtAl_CloudRoboticsCurrent_2016, SahaDasgupta_ComprehensiveSurveyRecent_2018, SiriweeraNaruse_SurveyCloudRobotics_2021}.

The interest in and relevance of software platforms for deploying robots at scale is also present in the commercial sector. Many private companies have started to build solutions for scaled deployment, fleet management, monitoring, teleoperation, and more. \textit{Balena} offers a Linux-based operating system, container engine, and cloud platform for deploying and managing massive fleets of IoT devices. While their deployment and monitoring systems are proprietary, the software components to be deployed are based on one or multiple composed Docker containers~\cite{_Balena_}. \textit{Formant} and \textit{Freedom Robotics} are other companies offering cloud platforms for robot fleet management, including ROS-based applications~\cite{_Formant_, _FreedomRobotics_}. Large players such as \textit{Amazon} have also built dedicated solutions: \textit{AWS IoT Greengrass} is an open-source edge runtime and cloud service for creating, deploying, and managing fleet devices and is also designed for ROS-based Docker containers~\cite{_AWSIoTGreengrass_}.

% ==============================================================================
\subsection{Automotive Service-Oriented Architectures}
\label{sec:automotive-service-oriented-architectures}

The development and deployment of highly-automated vehicles and ITS in general is one of the booming robotics fields, where large-scale deployment challenges arise. The paradigm shift towards software-defined vehicles~\cite{LiuEtAl_ImpactChallengesProspect_2022}, i.e., the value of vehicles is defined by software rather than hardware, is yet another example of "software is eating the world"~\cite{Andreessen_WhySoftwareEating_2011}. Similar to other industries, the trend towards ITS suggests to adopt a cloud-native approach~\cite{Shirasat_CloudnativeApproachSoftware_2021, CloudNativeComputingFoundation_CloudNativeDefinition_}. Automotive service-oriented or microservice architectures have been proposed and have been or are currently being (partially) implemented in various efforts in academia and industry~\cite{WagnerEtAl_AdaptiveSoftwareSystems_2011, FurstBechter_AUTOSARConnectedAutonomous_2016, WoopenEtAl_UNICARagilDisruptiveModular_2018a, RumezEtAl_OverviewAutomotiveServiceOriented_2020, PohnlEtAl_MiddlewareJourneyMicrocontrollers_, _ScalableOpenArchitecture_,LampeEtAl_RobotKubeOrchestratingLargeScale_2023}. These initiatives largely adopt cloud-native design patterns such as containerization, microservices, and orchestration.

Some parts of such architectures are already moving closer to production: \textit{Mercedes-Benz Research and Development North America} uses a microservice architecture relying on Kubernetes to fetch and push containerized app updates to connected vehicles~\cite{Microsoft_MercedesBenzCreatesContainerdriven_2020}. \textit{Volkswagen's CARIAD} uses Kubernetes for container-based simulation and testing of ECUs~\cite{RedHat_VolkswagenGroupBuilds_2021}.

% ==============================================================================
\subsection{Contributions}
\label{sec:contributions}

Considering the aforementioned trends and challenges, we recommend to facilitate, if not enable, the deployment of any-scale robotic applications, through the same techniques and tools that have proven successful in other domains: containerization and orchestration of microservices, supported by a backbone of automated DevOps processes. In particular, the contribution of this paper is three-fold:

\begin{itemize}
    \item We motivate, present, and discuss a novel microservice-based workflow for the development and deployment of robotic applications.
    \item We publicly release a tooling suite for enabling the proposed workflow, covering the automated minimal Docker containerization of ROS applications, a collection of useful machine learning-enabled base container images, and a Command Line Interface~(CLI) tool for simplified interaction with container images during the development phase.
    \item We qualitatively evaluate our tooling suite by comparing it to alternative Continuous Integration and Deployment~(CI/CD) solutions for robotic applications.
\end{itemize}

\section{Related Work}
\label{sec:related-work}

The automation of the software development life cycle, from building to testing to release, through DevOps processes such as CI/CD enables quick iterations and scalability, and integrates well with containerized microservice architectures. Our tooling suite falls exactly into this CI/CD category for ROS applications.

CI/CD has been an integral part of shipping ROS packages since the early 2010s. In 2015, the ROS build farm -- the official platform for building, testing, and releasing open-source ROS software as Debian packages -- adopted a containerized approach using Docker~\cite{DiMarco_DockerbasedROSBuild_2015, Thomas_ROSBuildFarm_2016}. While it is also possible to deploy a custom ROS build farm, its main purpose is to build, test, and release Debian packages, not container images of ROS packages.

\textit{ROS-Industrial} is a consortium dedicated to open-source industrial robotics and maintains the \textit{industrial\_ci} repository~\cite{ROS-Industrial_ROSIndustrial_, _IndustrialCi_2020}. \textit{industrial\_ci} contains CI configuration files for automating the build and test of ROS repositories. It supports ROS and ROS~2 distributions, integrates with popular CI platforms such as GitHub Actions or GitLab CI, and is largely configurable. While not being a core feature, it is also able to export the Docker image that is internally used for running the CI.

The \textit{ROS~2 Tooling Working Group} releases the \textit{setup-ros} and \textit{action-ros-ci} GitHub Actions~\cite{_GitHubActionSetup_2023, _GitHubActionROS_2023}. The provided CI configurations specifically targeted at GitHub can build and test ROS and ROS~2 workspaces. The CI processes can run in Docker containers, but are not restricted to them. The actions focus on the core parts of CI (build and test) and do not themselves deploy compiled ROS packages in any form.

Apart from the aforementioned projects, smaller tools have been released to cover some parts of the CI/CD process for ROS packages, but development has stalled or generic applicability is limited~\cite{Ferguson_BuildbotROS_2016, _ShadowRobotBuild_2018, _GitHubActionROS_2021, Lamoine_GitLabCIRos_2019}.

A popular base for the containerization of ROS applications is provided with the official ROS Docker images on DockerHub~\cite{OpenSourceRoboticsFoundation_DockerHubRos_}. Maintained by \textit{Open Robotics}, images for recent ROS distributions are provided in multiple variations with an increasing set of base dependencies. As of late 2023, the container images have been downloaded more than 10 million times, showcasing the popularity and relevance of containerizing ROS applications.

When it comes to deploying ROS applications with Docker, \textit{Canonical} advocates against using the de facto standard containerization tool. They claim that Docker comes with limitations with regards to robotics, and that these limitations are solved by their own edge device-oriented container solution named \textit{snaps}. They also present a case study of how Japanese robotics company \textit{Cyberdyne} deploys ROS software to their cleaning robots using \textit{snaps}. \cite{CanonicalLtd._ROSDockerWhy_2021, CanonicalLtd._KeepEnterpriseROS_2020}

\section{Overview of System Components}
\label{sec:overview-of-system-components}

Motivated by the trend towards microservice and service-oriented architectures of robotic applications, we recommend to follow the approach of microservice-based containerization and orchestration, built on top of automated DevOps processes. We present a container-driven workflow and tooling for developing and deploying ROS applications as Docker containers. This way, our proposed workflow and tooling suite act as enablers for the use cases, tools, and large-scale microservice architectures related to robotic applications as presented in~\sect{sec:introduction}. Due to its general applicability, widespread support, and popularity, we choose Docker as the containerization framework. We recognize \textit{Canonical's} criticism towards Docker in robotics as more of a tooling issue than a fundamental incompatibility, but still consider \textit{snaps} as an alternative way of deployment.

Our proposed workflow is illustrated and briefly explained in~\fig{fig:system-architecture}. The following sections will provide more details of its three main novel software components:
\begin{itemize}
    \item \textit{docker-ros}\footnote{\httpsurl{github.com/ika-rwth-aachen/docker-ros}} automatically builds minimal container images of ROS applications;
    \item \textit{docker-ros-ml-images}\footnote{\httpsurl{github.com/ika-rwth-aachen/docker-ros-ml-images}} provides machine learning-enabled base container images;
    \item \textit{docker-run}\footnote{\httpsurl{github.com/ika-rwth-aachen/docker-run}} is a CLI tool for simplified interaction with container images during the development phase.
\end{itemize}

% ==============================================================================
\subsection{Automated Container Image Build}
\label{sec:automated-deployment-image-build}

\begin{figure}[!t]
    \centering
    \smallskip
    \smallskip
    \includegraphics[width=\linewidth]{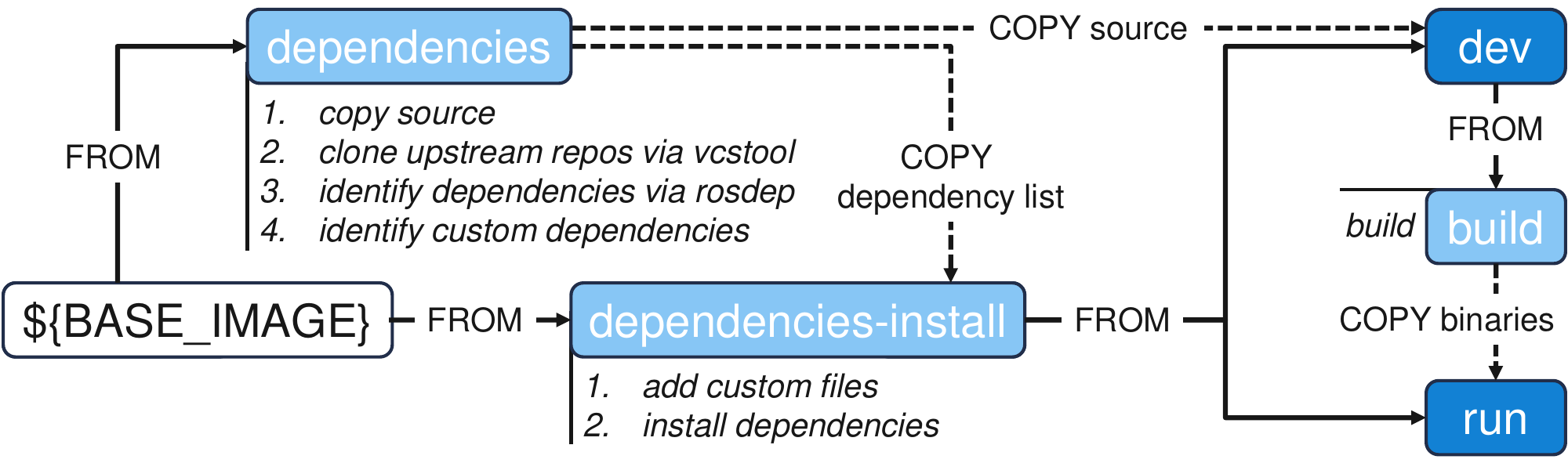}
    \setlength{\abovecaptionskip}{-15pt}
    \caption{Build stages of ROS containerization through generic Dockerfile}
    \label{fig:generic-dockerfile}
\end{figure}

In order to reduce containerization overhead, \textit{docker-ros} provides a generic multi-stage Dockerfile for building development and deployment Docker images of arbitrary ROS packages or package stacks. \fig{fig:generic-dockerfile} illustrates the image building process, including the sequence and interaction of individual build stages.

As base Docker image, any Ubuntu-based container image can be chosen. If the base image is missing a ROS core installation, it will be installed automatically. Dependencies of the to-be-containerized ROS application are preferably installed through ROS's built-in dependency management using \texttt{rosdep}. Dependencies on public and private code repositories are installed using \textit{vcstool}~\cite{Thomas_Vcstool_}. For a more complex dependency installation, one can also resort to custom shell scripts. All dependencies are identified in the \textit{dependencies} stage and subsequently installed in the \textit{dependencies-install} stage. The separation results in a stage that contains only the required dependencies without the associated target source code. Adding the source code again results in the development image (\textit{dev} stage). After building the target packages in a dedicated \textit{build} stage, only the built binaries (ROS install space) are copied to produce the minimal deployment image (\textit{run} stage).

The development image contains required all dependencies and the source code of the target ROS packages, while the deployment image only contains corresponding binaries and runtime dependencies. Apart from making the deployment image as lightweight as possible, this also allows to share the deployment image without disclosing source code.

We recommend to integrate the \textit{docker-ros} build process into a CI/CD pipeline for automated execution and container image builds. For this purpose, we provide template CI configurations for two of the most popular Git platforms: GitHub and GitLab. It is also possible to run the \textit{docker-ros} image building process locally. While the steps towards the container images are similar to what other ROS CI/CD tools already offer (e.g., \textit{industrial\_ci}), \textit{docker-ros} explicitly focuses on a simple, yet effective build of images specifically made for deployment as well as development.

% ==============================================================================
\subsection{Machine Learning-Enabled ROS Base Images}
\label{sec:machine-learning-enabled-ros-base-images}

The adoption of machine learning in robotic applications has gained traction in response to the increasing requirements of software components and recent successes in machine learning algorithms. For example, the development of highly-automated vehicles largely depends on data-driven machine learning algorithms for perception, planning, and more.

Therefore, with \textit{docker-ros-ml-images}, we provide a variety of lightweight multi-arch machine learning-enabled ROS Docker images. These serve well as base images for \textit{docker-ros} containerization, considering the growing application of data-driven algorithms in robotics. Each of the provided Docker images includes support for popular machine learning frameworks in addition to a ROS distribution. Currently, the supported machine learning frameworks are \textit{PyTorch}~\cite{PaszkeEtAl_PyTorchImperativeStyle_2019} and \textit{TensorFlow}~\cite{_TensorFlow_2023}. Since robotic applications are often implemented in C++ instead of Python for performance reasons, we also offer the C++ APIs of both frameworks. Combining the components listed in~\tab{tab:docker-ros-ml-images}, we have built more than 100 multi-arch images and make them publicly available on DockerHub. In addition to the provided images, we also publish the generic Dockerfile used to flexibly build images combining the different components.

\begin{table}[!t]
    \centering
    \smallskip
    \smallskip
    \caption{Components of ML-enabled ROS Docker Images}
    \label{tab:docker-ros-ml-images}
    \begin{tabular}{ll}
        \textbf{Component} & \textbf{Variations} \\
        \toprule
        ROS Distribution & noetic, foxy, humble, iron, rolling \\
        \rowcolor{gray!10}
        & \\
        \rowcolor{gray!10}
        \multirow{-2}{*}{ROS Components} & \multirow{-2}{*}{\shortstack[l]{core, base, robot, perception, \\desktop, desktop-full}} \\
        \multirow{2}{*}{ML Framework} & \multirow{2}{*}{\shortstack[l]{CUDA, TensorFlow Python, TensorFlow C++, \\PyTorch Python, PyTorch C++}} \\
        & \\
        \rowcolor{gray!10}
        Architecture & amd64, arm64 \\
        \bottomrule
    \end{tabular}
    \vspace{-7pt}
\end{table}

% ==============================================================================
\subsection{Modified Docker CLI with Useful Defaults}
\label{sec:modified-docker-cli-with-useful-defaults}

During development, it is advisable to employ the same environment as for later deployment. We suggest to adopt container-driven development using the development image built by \textit{docker-ros}. To enable a comfortable development workflow with Docker containers, we have built \textit{docker-run}.

\textit{docker-run} is a handy CLI tool based on the official Docker CLI, which not only starts a Docker container but also configures it to create optimal development conditions. Among other features, \textit{docker-run} automatically takes care of: X11~GUI forwarding from container to host; enabling GPU support; management of proper permissions inside the container; mounting existing source code into the container's ROS workspace; quickly attaching to a running container.

\textit{docker-run} is highly suitable for the workflow presented in \fig{fig:system-architecture}, but also works with Docker containers outside the realm of ROS and robotics. It is extensible through plugins, supports all flags of Docker's official CLI, and can therefore be used as an alternative to \texttt{docker run} / \texttt{exec}.

In the context of ROS, there is a similar tool maintained by \textit{Open Robotics}: \textit{rocker}~\cite{_Rocker_2023}. While also offering features like X11~GUI forwarding, \textit{rocker} relies on just-in-time Docker image builds and cannot be used for non-ROS purposes.

\section{Evaluation and Comparison}
\label{sec:qualitative-evaluation-and-comparison}

In this section, we evaluate the proposed containerization approach realized with \textit{docker-ros}. We first look at overhead compared to running processes natively on the host, then we compare \textit{docker-ros} to other existing CI/CD and/or containerization tools with regard to features relevant for automating robotic deployment. Finally, we present experiences gained by applying our methodology to our own automated vehicle.

% ==============================================================================
\subsection{Computational Overhead}
\label{sec:computational-overhead}

When it comes to adopting a new approach for deploying robotic software, runtime performance is critical for many robotic applications. Considering all types of computational overhead however, not only runtime performance, but also build time and disk size requirements are of interest.

% ------------------------------------------------------------------------------
\subsubsection{CI/CD and Build Time}
\label{sec:build-time}

Considering build time, the automated CI/CD process invoked by \textit{docker-ros} including containerization is mostly dominated by the involved dependency installation and source compilation steps. It will usually run asynchronously from development (e.g., on each repository push). A typical \textit{docker-ros} image building CI job takes between 2 and 10 minutes on modern hardware, mostly depending on the number and complexity of dependencies.

% ------------------------------------------------------------------------------
\subsubsection{Disk Space}
\label{sec:disk-space}

Microservices are characterized by minimal container images, i.e., they only contain the dependencies that are needed to run the particular application. In order to keep container image size on disk small, containerization tools like Docker reuse disk space of common base images, i.e., multiple \textit{docker-ros}-built images will reuse the disk space of the underlying ROS installation.

% ------------------------------------------------------------------------------
\subsubsection{Start-Up Time}
\label{sec:start-up-time}

Docker containers come with some start-up overhead compared to system processes. Exact overheads vary between different container runtimes like \textit{containerd}, \textit{CRI-O}, or \textit{runc}. In practice, start-up overheads below 1 s are measured~\cite{EspeEtAl_PerformanceEvaluationContainer_2020}, which we can also confirm with our own Docker container images built by \textit{docker-ros}. Note that this overhead scales with the number of containers that are launched simultaneously.

% ------------------------------------------------------------------------------
\subsubsection{Operation Performance}
\label{sec:operation-performance}

More important than the start-up overhead is to have \mbox{on-par} performance during normal operation of robotic software components, compared to running processes natively on the host system. Container technologies and Docker in particular have proven to only have negligible overhead for CPU, GPU, and memory performance~\cite{FelterEtAl_UpdatedPerformanceComparison_2015}.

% ==============================================================================
\subsection{Relevant Features for Automated Deployment}
\label{sec:relevant-features-for-automated-deployment}

As has been presented before, the deployment of any-scale robotic applications poses a set of challenges. In the following, we present the requirements and features that an automated build and containerization system for ROS applications -- such as we present with \textit{docker-ros} -- should fulfill. This will be the basis for the comparison in \sect{sec:feature-comparison-overview}.

\begin{table*}[!ht]
    \centering
    \smallskip
    \smallskip
    \setlength{\tabcolsep}{1pt}
    \caption{Feature comparison between ROS-related CI/CD and containerization tools as of \printdate{04.03.2024}.}
    \label{tab:comparision}
    \begin{threeparttable}
    \begin{tabular}{lccccc}
        \textbf{Feature} & \href{https://github.com/ika-rwth-aachen/docker-ros}{\textbf{\textit{docker-ros}}} & \href{https://github.com/ros-industrial/industrial_ci}{\textbf{\textit{industrial\_ci}}} & \href{https://github.com/ros-tooling/action-ros-ci}{\textbf{\textit{action-ros-ci}}} & \href{https://github.com/ros-infrastructure/ros_buildfarm}{\textbf{\textit{ros\_buildfarm}}} & \href{https://github.com/snapcore/action-build}{\textbf{\textit{snap}}} \\
        % \cmidrule{1-6}\morecmidrules\cmidrule{1-6}
        \toprule
        \arrayrulecolor{white}\midrule\arrayrulecolor{black}
        \multicolumn{6}{l}{\textbf{Compatibility and Versioning}} \\
        \midrule
        Any ROS / ROS~2 distribution\tnote{4}       & \cmark & \cmark & \cmark\tnote{5} & \cmark & \cmark \\
        \rowcolor{gray!10}
        Supported operating systems                 & Ubuntu & Ubuntu, Debian & Ubuntu, macOS, Windows & Ubuntu, Debian & Ubuntu \\
                                                    & & & & & \\
        \multirow{-2}{*}{Supported architectures}   & \multirow{-2}{*}{amd64, arm64} & \multirow{-2}{*}{amd64, arm64} & \multirow{-2}{*}{amd64} & \multirow{-2}{*}{\shortstack{amd64, arm64, \\armhf, i386}} & \multirow{-2}{*}{\shortstack{amd64, arm64, \\armhf, i386}} \\
        \rowcolor{gray!10}
        Build requirements                          & Docker & Docker & - & Docker & Snapcraft, LXD \\
        \arrayrulecolor{white}\midrule\arrayrulecolor{black}
        \multicolumn{6}{l}{\textbf{Source Dependency Management}} \\
        \midrule
        ROS distribution packages (\texttt{rosdep}) & \cmark & \cmark & \cmark & \cmark & \cmark \\
        \rowcolor{gray!10}
        System dependencies                         & \cmark & \cmark & \cmark\tnote{6} & \cmark & \cmark \\
        Pull of (private) upstream repositories     & \cmark & \cmark & \cmark & \cmark & \cmark \\
        \rowcolor{gray!10}
        Custom shell script                         & \cmark & \cmark & \cmark\tnote{6} & \xmark & \xmark \\
        \arrayrulecolor{white}\midrule\arrayrulecolor{black}
        \multicolumn{6}{l}{\textbf{Containerization / Packaging}} \\
        \midrule
        Build product                                       & Docker image & Docker image & \xmark & Debian package & snap \\
        \rowcolor{gray!10}
        Development image                                   & \cmark & \cmark & - & - & \xmark \\
        Minimal deployment image                            & \cmark & \xmark & - & - & \cmark \\
        \rowcolor{gray!10}
        Custom deployment launch command                    & \cmark & - & - & - & \cmark \\
        Custom base image                                   & \cmark & \cmark & - & - & \cmark \\
        \rowcolor{gray!10}
        Custom base image (without ROS)                     & \cmark & \cmark & - & - & \cmark \\
        Multi-architecture image                            & \cmark & \xmark & - & - & \cmark \\
        \rowcolor{gray!10}
        Runtime requirements                                & OCI runtime & OCI runtime & -  & package dependencies & snap \\
        \arrayrulecolor{white}\midrule\arrayrulecolor{black}
        \multicolumn{6}{l}{\textbf{Continuous Integration and Deployment (CI/CD)}} \\
        \midrule
        Target workspace build                          & \cmark & \cmark & \cmark & \cmark & \cmark \\
        \rowcolor{gray!10}
        Downstream workspace build                      & \cmark\tnote{7} & \cmark & \xmark & \cmark & \xmark \\
        Testing                                         & \cmark\tnote{7} & \cmark & \cmark & \cmark & \xmark \\
        \rowcolor{gray!10}
        Containerization/Packaging                      & \cmark & \cmark & \xmark & \cmark & \cmark \\
        Generation of documentation                     & \xmark & \xmark & \xmark & \cmark & \xmark \\
        \rowcolor{gray!10}
                                                        &  &  &  &  & \\
        \rowcolor{gray!10}
                                                        &  &  &  &  & \\
        \rowcolor{gray!10}
        \multirow{-3}{*}{Supported CI/CD platforms}     & \multirow{-3}{*}{\shortstack{GitHub Actions, \\GitLab CI}} & \multirow{-3}{*}{\shortstack{GitHub Actions, GitLab CI, \\Travis CI, Bitbucket Pipelines, \\Google Cloud Build}} & \multirow{-3}{*}{GitHub Actions} & \multirow{-3}{*}{\shortstack{GitHub Actions, \\Jenkins}} & \multirow{-3}{*}{GitHub Actions} \\
        \arrayrulecolor{white}\midrule\arrayrulecolor{black}
        \multicolumn{6}{l}{\textbf{Maturity and Popularity}} \\
        \midrule
        Last updated (main branch)  & \printdate{16.02.2024} & \printdate{26.01.2024} & \printdate{24.01.2024} & \printdate{31.01.2024} & \printdate{27.10.2023} \\
        \rowcolor{gray!10}
        Last release                & \printdate{31.01.2024} & \printdate{08.04.2020} & \printdate{24.01.2024} & \printdate{11.06.2019} & \printdate{24.10.2023} \\
        First release               & \printdate{28.05.2023} & \printdate{09.12.2015} & \printdate{18.10.2019} & \printdate{02.02.2016} & \printdate{09.04.2020} \\
        \rowcolor{gray!10}
        Contributors                & 8 & 35 & 26 & 35 & 8 \\
        GitHub stars                & 100 & 227 & 132 & 75 & 37 \\
        \arrayrulecolor{white}\midrule\arrayrulecolor{black}
        \bottomrule
    \end{tabular}
    \begin{tablenotes}
        \item[4] Here, \textit{any} refers to the currently supported ROS distributions, but it does not necessarily imply that a specific EOL distribution is not supported.
        \item[5] ROS~2's default build tool \texttt{colcon} is also used to build ROS 1 packages instead of \texttt{catkin}, which might cause problems in some cases.
        \item[6] GitHub action CI configuration file allows to specify custom shell commands for dependency installation, independent of \textit{action-ros-ci}.
        \item[7] Realized by integrating \textit{industrial\_ci} as testing stage into the CI configurations provided by \textit{docker-ros}.
    \end{tablenotes}
    \end{threeparttable}
    \vspace{-15pt}
\end{table*}

% ------------------------------------------------------------------------------
\subsubsection{Compatibility and Versioning}
\label{sec:compatibility-and-versioning}

While the second major ROS version (ROS~2) has been officially released since 2017, the original ROS version is still widely used today, mainly due to its richer ecosystem of existing packages. These two major ROS versions each have subversions that are regularly released as so-called ROS distributions. A general-purpose ROS deployment tool should support as many ROS versions and distributions as possible.

With support for any ROS version and distribution comes also the need to support different operating systems. The heterogeneity of computing devices present in large-scale systems such as ITS also necessitates supporting different computing architectures. ROS and ROS~2 mainly target the Ubuntu operating system on amd64 and arm64 architectures~(tier~1 support)~\cite{FooteConley_ROSREPTarget_2010, ArguedasEtAl_ROSREP2000_2018}. ROS~2 also considers Windows 10 on the amd64 architecture a tier~1 platform. Other operating systems such as Debian or macOS are only considered tier~2, i.e., they are not thoroughly tested and must be built from source, but are reported to be functional.

In terms of dependencies, the containerization approach requires devices to run a container runtime, in our case the Docker Engine. Docker itself is supported on any of ROS's tier~1 operating systems and architectures.

% ------------------------------------------------------------------------------
\subsubsection{Source Dependency Management}
\label{sec:source-dependency-management}

ROS nodes may be written in C++ or Python and often depend on other ROS packages. The preferred way for specifying and installing such dependencies is \textit{rosdep}, which can also automatically install certain system dependencies.

Some system dependencies may not be available through \textit{rosdep}. In such cases, developers often resort to custom shell scripts for the advanced installation of dependencies. An automated build system should therefore provide a way of falling back to custom shell script execution.

Apart from system and ROS distribution dependencies, ROS source code may also depend on other (potentially privately hosted) source code that is included as a Git submodule or via other version control tools (e.g., \textit{vcstool}).

% ------------------------------------------------------------------------------
\subsubsection{Containerization/Packaging}
\label{sec:containerization}

Deploying ROS packages in the form of minimal containers has been motivated before. This also enables the distribution of the image without disclosing source code. Since the purpose of the deployment image is to run the containerized function, the container launch command has to be configurable to directly start the function with default configuration.

Before deployment, it makes sense to develop a given ROS package within the same environment as it is later being deployed to. This can be achieved by a container-driven development workflow as described before.

While an automated containerization platform should allow to integrate arbitrarily complex dependency installation, some core system dependencies are much easier to integrate by specifying a suitable container base image.

With regard to general support for different computing architectures, containerization tools like Docker also support multi-architecture images. This makes deployment to different architectures easier, as the same image names can be used across architectures.

% ------------------------------------------------------------------------------
\subsubsection{Continuous Integration and Deployment~(CI/CD)}
\label{sec:continuous-integration-and-deployment}

A large driver in speeding up software life cycles is the adoption of DevOps techniques such as Continuous Integration and Deployment~(CI/CD). In the context of ROS-based applications, this involves the automation of the following on changes to the code base: dependencies of the ROS source packages are installed; the ROS source is built; the ROS source is installed; tests are executed, if implemented; documentation is generated, if present; and finally, build products (binaries or Python scripts) may be deployed to deployment targets. In order to ensure a controlled environment and reproducible CI runs, modern CI/CD software oftentimes relies on running in isolated containers.

Building and testing in the CI workflow should consider upstream, target, and downstream ROS workspaces. Upstream packages are needed for building and testing the target or downstream packages. Downstream packages depend on the target workspace and should get tested against it.

There are many established CI/CD platforms to choose from, e.g., \textit{GitHub Actions} or \textit{GitLab CI}. Tools for realizing the ROS-specific CI/CD workflow should ideally integrate well with existing ones at a particular organization.

% ------------------------------------------------------------------------------
\subsubsection{Maturity and Popularity}
\label{sec:maturity-and-popularity}

One important aspect to consider when evaluating different CI/CD frameworks for ROS is the maturity and popularity of the framework. In today's fast-moving software world, such a framework usually loses value as soon as it is not being maintained anymore. Apparent popularity (e.g., in the form of GitHub stars) may also be a good indicator of the framework's quality and value it is providing to developers.

% ==============================================================================
\subsection{Feature Comparison Overview}
\label{sec:feature-comparison-overview}

With regard to the previously presented criteria, \tab{tab:comparision} compares \textit{docker-ros}, \textit{industrial\_ci}, \textit{action-ros-ci}, \textit{ros\_buildfarm}, and \textit{snap} to each other. Besides \textit{docker-ros}, the four other tools are also associated with continuous integration for ROS applications and already briefly introduced in Section~\ref{sec:related-work}. The table is separated into the five categories mentioned in Section~\ref{sec:relevant-features-for-automated-deployment}.

The comparison highlights that \textit{industrial\_ci} and \textit{action-ros-ci} are not designed for the containerization of ROS applications, but instead focus on running CI. In the case of \textit{ros\_buildfarm}, producing Debian packages is a valid method for software distribution, but comes short of integrating well into microservice architectures as motivated in \sect{sec:introduction}. The \textit{snap} method comes closest to the containerized approach, but also lacks applicability in large-scale orchestrated software architectures. Another shortcoming of \textit{snaps} is that they cannot be used to create a development environment.

While also integrating core CI elements, \textit{docker-ros} focuses on building minimal Docker images for container-driven development and deployment of ROS applications. The automatically built container images can directly be deployed using large-scale container management and orchestration frameworks like \textit{docker-compose}, \textit{docker-swarm}, or \textit{Kubernetes}.

% ==============================================================================
\subsection{The Container-Driven Workflow in Practice}
\label{sec:the-container-driven-workflow-in-practice}

\begin{figure}[!t]
    \centering
    \smallskip
    \smallskip
    \includegraphics[width=\linewidth]{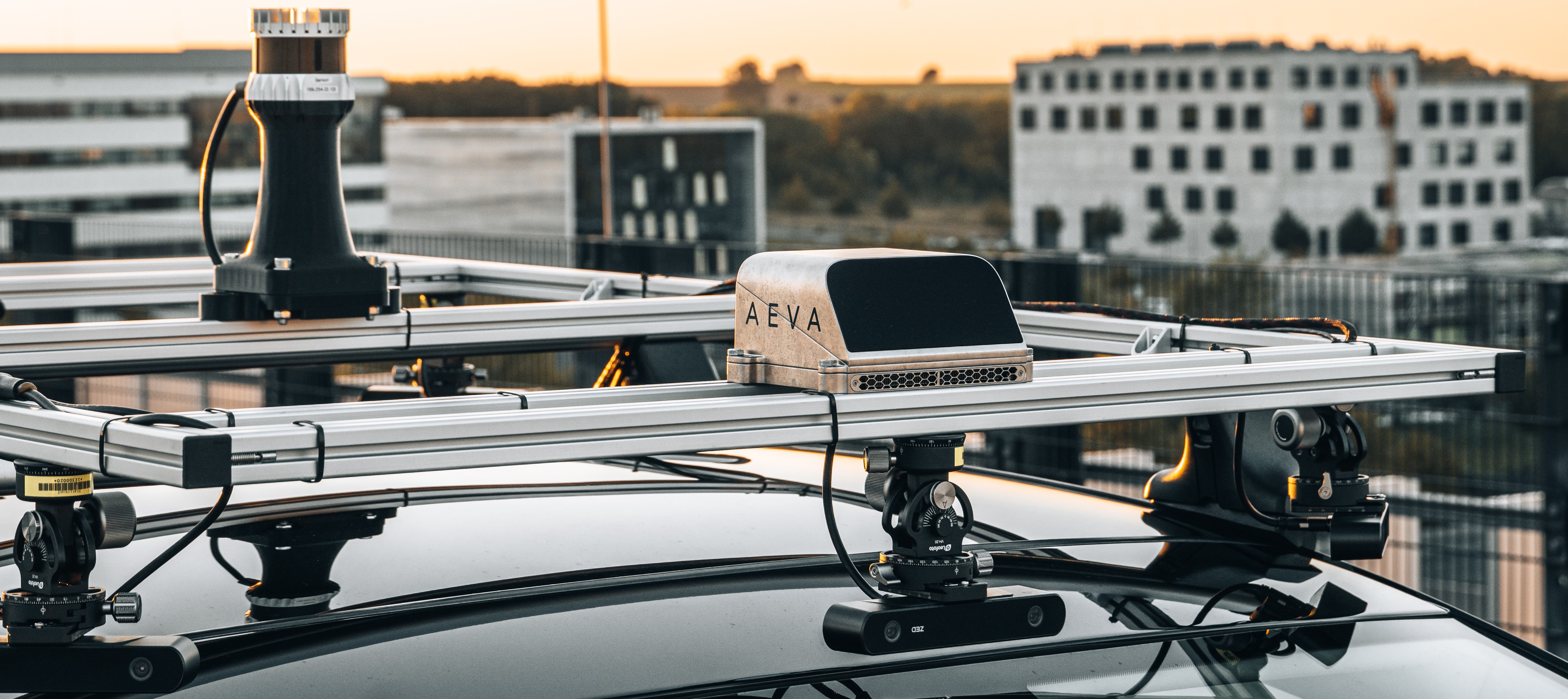}
    \setlength{\abovecaptionskip}{-15pt}
    \caption{Sensor rack of research vehicle; as the rest of the software stack, the drivers for the sensors are each running in their own minimal container.}
    \label{fig:rack}
\end{figure}

Apart from the qualitative evaluation and comparison of our proposed containerization approach, we also present experiences made by applying the methodology to our existing automated driving software stack and deployment of the stack to our research vehicle, see~\fig{fig:rack}.

The software stack is composed of many individual modular ROS packages, managed in dedicated Git repositories. Through the \textit{docker-ros} GitLab CI integration, all packages are automatically built, tested, and uploaded as minimal development and deployment container images to our container registry. For deployment in the research vehicle, \textit{docker-compose} compositions of individual container images are launched. No source code is built or even present on the vehicle, unless a specific module requires in-place development, assisted through \textit{docker-run}.

Adopting the proposed workflow to our automated driving stack and vehicle testbed has shown several advantages: acceleration of development cycles through decoupling individual modules; simple and consistent integration of automated CI/CD processes for all modules; availability of pre-built container images for all modules in order to focus on deployment-related tasks; and readiness for large-scale cooperative multi-robot systems as presented in~\cite{LampeEtAl_RobotKubeOrchestratingLargeScale_2023}.

\section{Conclusion}
\label{sec:conclusion}

We have motivated, presented, and discussed a novel microservice-based workflow dedicated to enabling the deployment of any-scale robotic applications. The proposed workflow revolves around a novel tooling suite that is publicly released as part of this work. The tooling suite consists of \textit{docker-ros} for simple automated container image builds, \textit{docker-ros-ml-images} for machine learning-enabled ROS base images, and \textit{docker-run} for an easy container-driven development workflow on the command line. All tools are made open-source and free to use in order to spread their usage throughout the ROS and Docker communities.

In addition to embedding the workflow into the overall context of robotic software deployment, we have also evaluated our tooling suite by comparing it to alternative solutions. Our in-depth feature analysis of prominent ROS CI/CD tools has found value in all existing solutions, but has highlighted \textit{docker-ros's} focus on building deployment container images.

Finally, we have presented first-hand experiences made by adopting the proposed workflow for our automated driving software stack, highlighting the advantages that the workflow brings to developers, system integrators, and other stakeholders in the robotic software deployment process.

%-------------------------------------------------------------------------------

% This command serves to balance the column lengths
% on the last page of the document manually. It shortens
% the textheight of the last page by a suitable amount.
% This command does not take effect until the next page
% so it should come on the page before the last. Make
% sure that you do not shorten the textheight too much.
\addtolength{\textheight}{-4.5cm}

%-------------------------------------------------------------------------------

\clearpage

% dont print URL in references
%\renewcommand{\url}[1]{\href{#1}{Link}}

\bibliographystyle{IEEEtran}
\bibliography{references}

\end{document}